\documentclass[letterpaper, 10 pt, conference]{ieeeconf}  

\IEEEoverridecommandlockouts                              
\usepackage{verbatim}
\usepackage{mathtools}
\usepackage{amssymb}
\usepackage{enumerate}
\usepackage{tikz}
\usepackage{tikzscale}
\usepackage[hidelinks]{hyperref}
\usepackage[multiple]{footmisc}
\usetikzlibrary{positioning}
\usetikzlibrary{shapes.geometric, arrows}

\tikzstyle{startstop} = [rectangle, rounded corners, minimum width=3cm, minimum height=1cm,text centered, draw=black, fill=red!30]
\tikzstyle{io} = [trapezium, trapezium left angle=70, trapezium right angle=110, minimum width=0cm, minimum height=0cm, text centered, draw=black, fill=blue!30]
\tikzstyle{process} = [rectangle, minimum width=0cm, minimum height=0cm, text centered, text width=3cm, draw=black, fill=orange!30]
\tikzstyle{decision} = [diamond, minimum width=0cm, minimum height=0cm, text centered, draw=black, fill=green!30]
\tikzstyle{arrow} = [thick,->,>=stealth]

\title{\LARGE \bf
Map Management for Efficient Long-Term Visual Localization in Outdoor Environments
}

\author{Mathias B\"{u}rki, Marcin Dymczyk, Igor Gilitschenski, Cesar Cadena, Roland Siegwart, and Juan Nieto
\\ Autonomous Systems Lab, ETH Z\"{u}rich
\\ {\tt\footnotesize \{firstname.lastname\}@mavt.ethz.ch}
}

\newcommand{\pastobs}{O}

\newcommand{\appearance}{\mathcal{A}}
\newcommand{\selected}{S}

\newcommand{\allsessions}{Z}

\newcommand{\allcandidates}{C}

\newcommand{\allselected}{S}
\newcommand{\alllandmarks}{L}
\newcommand{\landmark}{l}

\newcommand{\card}[1]{\vert #1 \vert}

\newcommand{\selectionratio}{sr}

\newcommand{\observationsession}{\textit{observation session}}
\newcommand{\observationsessions}{\textit{observation sessions}}
\newcommand{\richsession}{\textit{rich session}}
\newcommand{\richsessions}{\textit{rich sessions}}
\newcommand{\readability}{\textit{Note: Improved readbility on computer screen.}}

\begin{document}

\maketitle
\thispagestyle{empty}
\pagestyle{empty}

\begin{abstract}

We present a complete map management process for a visual localization system designed for multi-vehicle long-term operations in resource constrained outdoor environments.
Outdoor visual localization generates large amounts of data that need to be incorporated into a lifelong visual map in order to allow localization at all times and under all appearance conditions. 
Processing these large quantities of data is non-trivial, as it is subject to limited computational and storage capabilities both on the vehicle and on the mapping backend.
%
We address this problem with a two-fold map update paradigm capable of, either, adding new visual cues to the map, or updating co-observation statistics.
The former, in combination with offline map summarization techniques, allows enhancing the appearance coverage of the lifelong map while keeping the map size limited.
On the other hand, the latter is able to significantly boost the appearance-based landmark selection for efficient online localization without incurring any additional computational or storage burden.
%
%
%
Our evaluation in challenging outdoor conditions shows that our proposed map management process allows building and maintaining maps for precise visual localization over long time spans in a tractable and scalable fashion.


\end{abstract}

\section{Introduction}


Visual localization systems for mobile robots constitute an attractive alternative to laser-based systems, as the former can offer accurate localization performance with a low-cost sensor setup. 
This is especially true for future autonomous cars, where mass-production renders the sensor suite a sensitive matter of expense. 
However, visual localization systems generate large amounts of data that need to be processed both online on the vehicles as well as offline through map building and map maintenance.
The inherent sensitivity of visual systems with respect to changing appearance conditions further exacerbates this problem in the context of long-term autonomy, as multiple appearances of the mapped places need to be stored and managed in order to be able to localize with satisfying accuracy across all conditions. 

\begin{figure}
\includegraphics[width=0.5\textwidth]{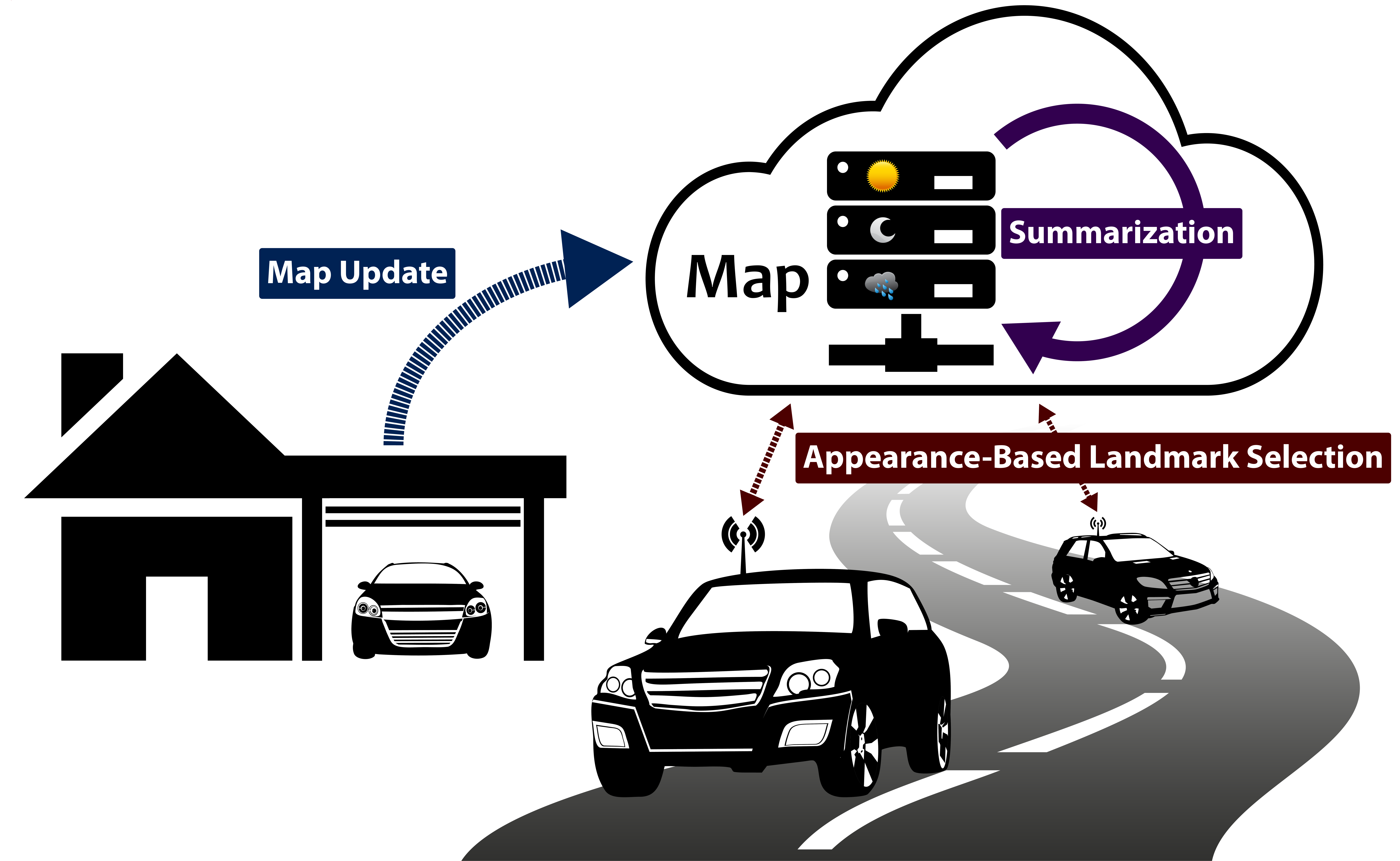}
\caption{The model scenario motivating this work. 
Multiple vehicles simultaneously localize using a shared remote map, which is accessed over a mobile communication link. 
Appearance-based landmark selection allows only querying landmarks from the map which are in accordance with the current appearance condition, ensuring efficient usage of the available bandwidth.
Once a vehicle returns from a sortie, collected map data is uploaded to the backend and incorporated into the map. 
Subsequent map summarization on the backend ensures the map size never exceeds a fixed number of landmarks, and thus guarantees limited storage requirements and a tractable map maintenance process in long-term perspective.
\label{fig:scenario_front}}
\vspace{-5mm}
\end{figure}

%
%
In recent years, methods have been presented to address the scalability and efficiency issues of individual components of a visual localization system (\cite{Muhlfellner2015SummaryLocalization, Dymczyk2015KeepMaps, Burki2016Appearance-basedLocalization, Dymczyk2016ErasingMapping, Churchill2013Experience-basedLocalisation, Paton2016BridgingRepeat}).
However, little attention has been paid to how to combine the different components to form a complete and tractable localization framework, how the differently optimized methods interact, how visual maps are to be built and managed over indefinite time spans, and most importantly: how the large amount of data accumulated over time can be processed and incorporated in an optimal way; everything with the purpose of allowing precise visual localization in outdoor environments at all times.
It is the aim of this paper to address these questions.
For this, we have built a scalable and efficient visual localization system for multi-vehicle outdoor shared-map scenarios as depicted in Figure~\ref{fig:scenario_front} and anticipated for autonomous cars in the near future.
It employs both appearance-based landmark selection on the vehicle side, as well as offline map summarization on the cloud-based mapping backend.
We demonstrate how the visual maps can be managed and improved over time as the vehicles are exposed to vastly different appearance conditions. 
With the novel concept of an $\observationsession$, together with a modified formulation of the ranking function for appearance-based landmark selection, we propose a lightweight procedure to handle large quantities of frequently collected sensor data with the aim of improving the landmark selection performance without increasing the size of the map.


%

Our main contributions can be summarized as follows:
\begin{enumerate}
\item Demonstration of a complete map management procedure for an efficient visual localization and mapping system designed for long-term outdoor use. 
\newpage
\item Introduction of $\observationsessions$ and proposal of a new formulation of the ranking function for appearance-based landmark selection, allowing to exploit frequently collected sensor data to significantly increase the landmark selection performance without increasing the map size.

\end{enumerate}

In an extensive evaluation in two real-world scenarios, covering weather and seasonal changes at daylight over the course of one year, and the extreme illumination change from day-time to night-time over the course of one day, we validate, first, the practicability of the proposed map management procedure in challenging outdoor conditions, and second, we show how additional co-observability statistics can improve the appearance-based online localization, and where the limitations thereof lie. 

The rest of this paper is structured as follows:
After an overview over related literature, we present our map management procedure in detail in Section~\ref{sec:methodology}, before presenting an extensive evaluation of the system's performance in Section~\ref{sec:evaluation}.
Summarizing remarks about our key findings conclude the paper in Section~\ref{sec:conclusions}.
\section{related work}

Ever since the advent of SLAM systems, maintaining map representations for enabling long-term operations has been a key focus, with a variety of different approaches evolving over time. 
The methods described in \cite{Walcott-Bryant2012DynamicEnvironments}, \cite{Dayoub2011Long-termEnvironments} and  \cite{Dayoub2008AnRobots} aim at maintaining a most up-to-date representation of the environment over longer times.
These approaches, however, reach their limits whenever a map is required to represent multiple different representations at the same time, as is the case for outdoor visual localization applications.

For this reason, substantial efforts have been made to permanently augment visual maps with data from differing environment conditions.
Churchill et al. \cite{Churchill2013Experience-basedLocalisation} introduced the Experience-Based mapping framework, which on-demand adds new ``sub-maps'' (referred to as ``Experiences'') of the environment under newly observed appearance conditions.
In a similar vein, the Multi-Session mapping proposed by M{\"u}hlfellner et al. \cite{Muhlfellner2015SummaryLocalization} and the Multi-Experience Localization proposed by Paton et al. \cite{Paton2016BridgingRepeat} both allow adding multiple datasets of an environment, collected during differing appearance conditions, to a common map representation.
All of these approaches in their basic form, however, suffer from increased and ultimately unbounded storage, memory and computational resource requirements.
To address this deficit, map summarization techniques have been developed, aiming at maintaining as small a map representation as possible, while at the same time still providing as good a localization performance across as far ranging appearance conditions as possible. 
Early works in this field include identifying reliable, geometrically-consistent features in the image retrieval context~\cite{Turcot2009BetterProblems}, and suppressing confusing features from certain regions of the database images~\cite{Knopp2010AvoidingRecognition}.
Follow-up approaches vary from clustering \cite{Konolige2009TowardsMaps} or random pruning \cite{Milford2010PersistentSystem} of visual ``Views'', to selection at the landmarks level based on various landmark ranking functions \cite{Muhlfellner2015SummaryLocalization}, \cite{Dymczyk2015KeepMaps}, \cite{Dymczyk2015TheLocalization}, \cite{Pirker2011CDWorld}, \cite{loquercio2017efficient}.

The selection must not necessarily be carried out on the backend side, but instead may as well be performed already in an online fashion on the robot, prior to uploading new map data \cite{Hartmann2014PredictingMatch, Dymczyk2016ErasingMapping, Rosen2016TowardsEnvironments, Rosen2016TowardsEnvironments}.
%
%
In general, these contributions focus on constructing a reliable set of landmarks for all possible environment states, reducing the runtime of tracking and localization, and/or the uplink bandwidth requirements.

%
In contrast to that, online landmark selection algorithms further allow decreasing the resource demands on the vehicle and on the communication downlink by having the vehicle query the map only for a selective fraction of landmarks which are deemed useful under the current operating conditions.
%
Previous work by the authors \cite{Burki2016Appearance-basedLocalization} and by Linegar et al. \cite{Linegar2015WorkLocalisation} have successfully demonstrated such algorithms in the context of Autonomous Driving -- a use-case especially prone to visual appearance change and applicable to the shared-map scenario.
In relation to that, the work by Krajn\'ik \cite{Krajnik2016PersistentEnhancement}, \cite{Krajnik2014Long-termMaps} aims at predicting the current state of the environment based on previously observed and learned temporal patterns.
While this approach is promising for dynamic indoor applications, it is only partially applicable to outdoor environments with often non-periodic changes.

We believe that an ultimately efficient visual localization system must do justice to constrained resources along the whole pipeline, that is, on the mapping backend side, as well as on the mobile platform and the communication link in-between. 
None of the aforementioned works, however, address all of these constraints simultaneously, whereas in this paper, we present a map management procedure that allows reaching an entirely scalable and efficient visual localization system for long-term use.
%
%
%
Furthermore, and in contrast to \cite{Linegar2015WorkLocalisation} and \cite{Paton2016BridgingRepeat}, our metric multi-session map representation (see \cite{Muhlfellner2015SummaryLocalization}, \cite{Dymczyk2015TheLocalization}) keeps all map data (vertices, landmarks), even from multiple appearance conditions, expressed in a single map reference frame.
This not only facilitates higher level tasks of autonomous operation, such as path planning and control, but also allows implementing the online landmark-selection and the offline map summarization on the level of individual landmarks.





\section{Methodology}
\label{sec:methodology}
In this section, we present the theoretical concepts of the three main components of our localization and mapping system: (i) the map update, (ii) appearance-based landmark selection with $\observationsessions$, and offline map summarization.
\subsection{Map Update}
\label{subsec:map_update}
\begin{figure}
\centering
\includegraphics[width=0.5\textwidth]{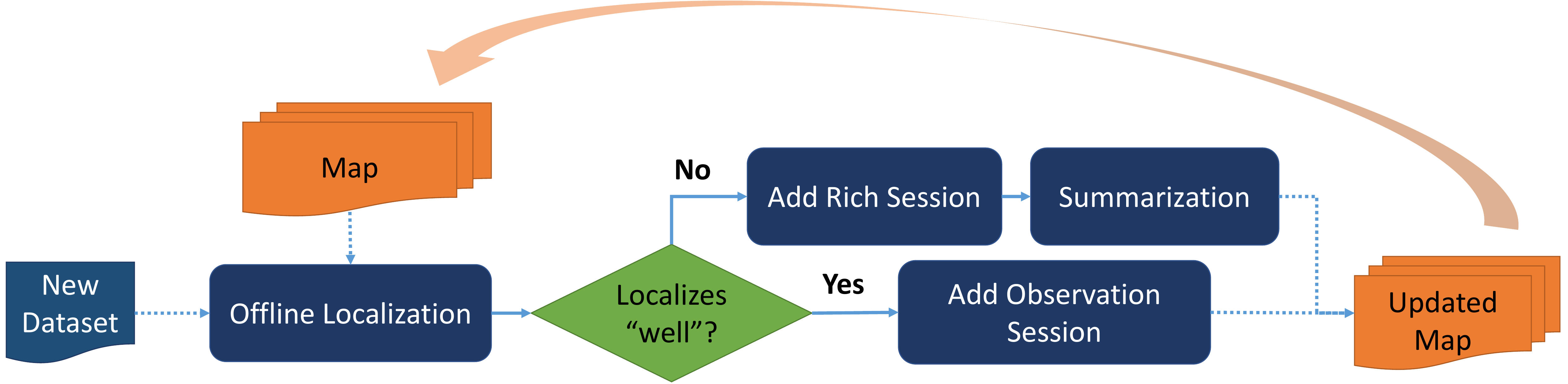}
\caption{Schematic illustration of the map update process on the backend.
A~newly received dataset is first localized against the available map. In case the localization performance is below a defined threshold, new landmarks are created from the dataset and added to the map, followed by a summarization step, which reduces the total number of landmarks in the map again to a fixed number. In the other case, the co-observation statistics of all landmarks observed during the localization are updated, but no new landmarks are added.
\label{fig:flowchart_map_update}}
\vspace{-4mm}
\end{figure}

The methodology of our map management procedure is based on a map update process as depicted in Figure~\ref{fig:flowchart_map_update}.
Sensor data, consisting of camera images and wheel-odometry measurements, is collected during a sortie of a vehicle and processed after the vehicle has returned to its home-base. 
The newly collected dataset is first localized in an offline process against the map available at that time.
In case the performance of this localization is worse than a pre-defined threshold, the map is considered to not cover the appearance condition encountered during this sortie sufficiently well and new landmarks are tracked and triangulated from the dataset. 
A dataset added to the map in this fashion is referred to as a $\richsession$.
A subsequent map summarization step ensures the total number of landmarks to remain below a fixed number, guaranteeing a bounded map size at all times.
If, on the other hand, localization has performed sufficiently well, the map is considered to cover the encountered conditions and no new landmarks are added to the map. 
In this case, however, the localization still reveals useful information about what landmarks in the map have been observed during the sortie.
This information is added to the map in the form of an $\observationsession$.
In contrast to the $\richsession$ described above, adding an $\observationsession$ conforms to merely marking existing landmarks as observed in the respective sortie.
%
However, both for future online localization as well as future map summarization steps, this additional statistical data is valuable, as it allows a better distinction between useful and not useful landmarks.
The resulting updated map is then used for localization of subsequent sorties.

In order to benefit from the $\observationsessions$ during online localization with appearance-based landmark selection, a modified formulation of the landmark ranking function is required and described in the following subsection.  


\subsection{Appearance-Based Landmark Selection with Observation Sessions}
\label{subsec:methodology_abl}
In our previous work presented in \cite{Burki2016Appearance-basedLocalization}, a method to tackle the problem of only querying useful map data during an outdoor operation has been introduced on the basis of appearance-based landmark selection.
%
Following an iterative localization paradigm, such as the ones described in \cite{Muhlfellner2015SummaryLocalization} and \cite{Lategahn2012CityVision}, a ranking function $f(\landmark)$ assigns a score to each landmark of a candidate set $\allcandidates_k$ (pre-selected based on spatial proximity), according to how likely $\landmark$ is observable under the current appearance condition.
Then, a small subset $\allselected_k$ of top-ranked landmarks are selected using a selection policy $\Omega(f(), \ldots)$, transmitted to the vehicle, and used for localization at iteration $k$.
The ranking function described in \cite{Burki2016Appearance-basedLocalization} adaptively weights the different sessions present in the pre-built map based on the session-affiliation of recently observed landmarks along the traversal.
Although successfully reducing the amount of landmarks used for localization, it relies on the map to be created a priori with all sessions approximately uniformly distributed across the appearance space.

In a practical scenario, however, the map sessions may not be uniformly distributed, but they are rather added once a ``new'' appearance condition is encountered for the first time.
In addition to that, whenever the vehicle traverses through the mapped area under an appearance condition already well-covered in the map, additional co-observability information can be gathered in the form of $\observationsessions$.
%
%
%
%

As our experiments presented later show (see Figure~\ref{fig:zoo_fofa_ncm_bars}), the original ranking function from \cite{Burki2016Appearance-basedLocalization}, denoted by $f_{orig}()$, is not well suited to incorporate these additional $\observationsessions$. 
We thus propose a new formulation of the ranking function that is agnostic to how the mapping sessions are distributed across the appearance space, and the number and distribution of additional $\observationsessions$ present in the map.

Let $\appearance$ denote the current appearance condition. 
We are interested in evaluating $p(\landmark \mid \appearance)$, corresponding to the probability of observing landmark $\landmark$ under the current appearance condition.
Let further $\allsessions$ denote the set of all sessions present in the map, both $\richsessions$ and $\observationsessions$, and $\alllandmarks$ denote the set of all landmarks in the map.
With every landmark $\landmark \in \alllandmarks$, we associate the set $\allsessions_l$, corresponding to all sessions that have observed landmark $\landmark$.

We note that $p(\landmark \mid \appearance)$ directly depends on $\allsessions_l$, that is, $\allsessions_{l_i} = \allsessions_{l_j} \Rightarrow p(\landmark_i \mid \appearance) = p(\landmark_j \mid \appearance)$.
We can thus group landmarks into appearance equivalence classes, according to:
\begin{equation}
[\landmark_i] \coloneqq \lbrace \landmark_j \in \alllandmarks \mid \allsessions_{l_j} = \allsessions_{l_i} \rbrace
\end{equation}
with
\begin{equation}
p(\landmark_j \mid \appearance) = p(\landmark_i \mid \appearance) \text{ } \forall \landmark_j \in [\landmark_i]
\end{equation}

Hence, evaluating ${p(\landmark \mid \appearance)}$ amounts to evaluating ${p([\landmark] \mid \appearance)}$, which can be interpreted as the relevance of appearance equivalence class $[\landmark]$ under appearance condition\,$\appearance$.

\begin{figure}
\includegraphics[width=0.5\textwidth]{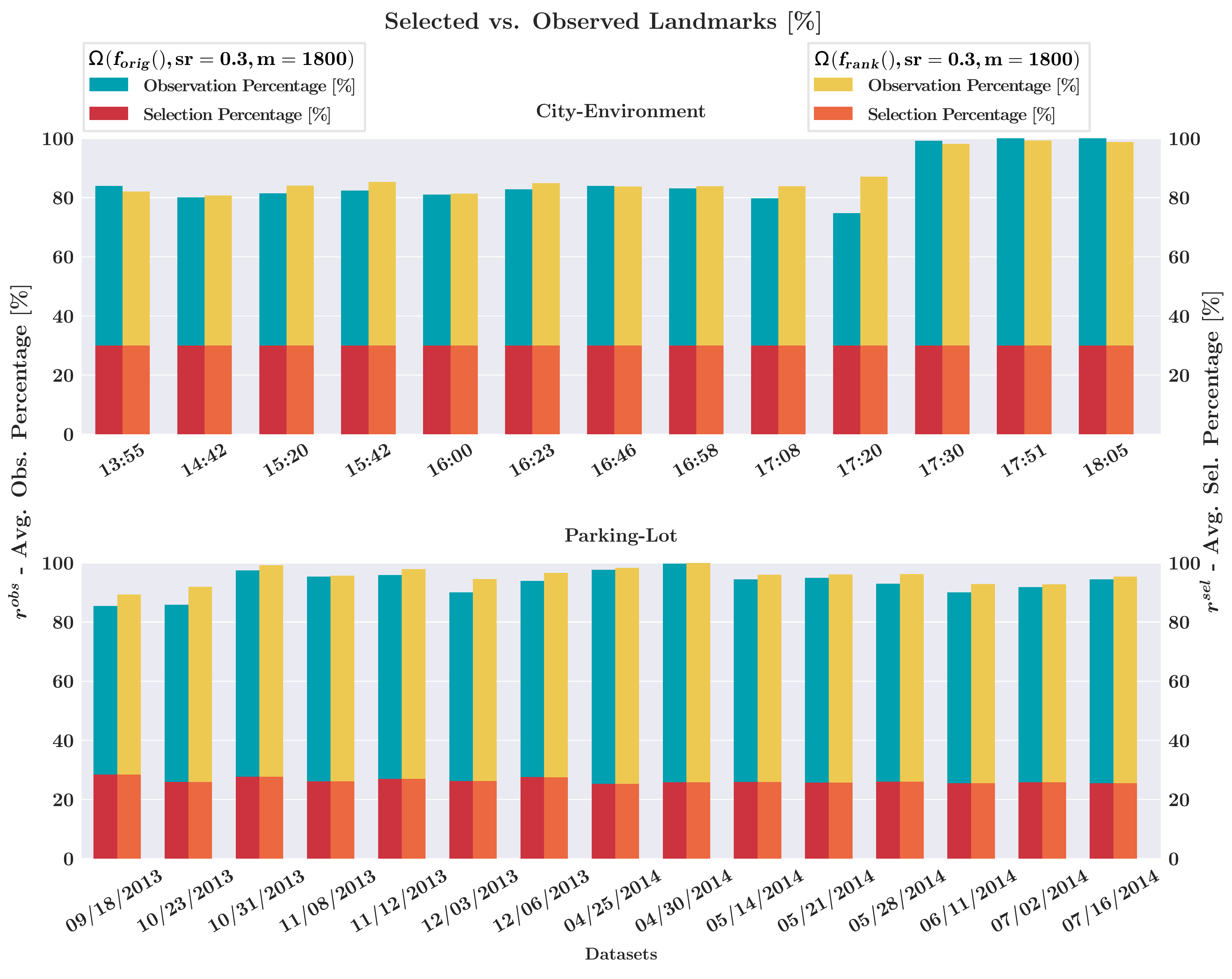}
\caption{Direct comparison of our new appearance-based ranking function ${f_{rank}()}$ with the original ranking function proposed in \cite{Burki2016Appearance-basedLocalization} ${f_{orig}()}$, ensuring the former to perform at least as well on the same map and selection policy used for the experiments in \cite{Burki2016Appearance-basedLocalization}.
For each dataset of the City-Environment and the Parking-Lot scenario, the observation percentage $r^{obs}$ (blue, yellow bars) is shown for a selection ratio of $30\%$ and a maximum number of selected landmarks of $m=1800$. 
In this setting (a priori built map without $\observationsessions$) both ranking functions perform equally well. The benefit of our modified ranking function ${f_{orig}()}$ is shown in further experiments with $\observationsessions$ presented in section \ref{subsec:evaluation_abl}. 
\label{fig:iros2016_comp}}
\vspace{-4mm}
\end{figure}

The abstract appearance condition $\appearance$ is not directly observable.
However, it can be approximated by the means of recently selected and observed landmarks as follows:
\begin{equation}
p([\landmark_i] \mid \appearance) \approx
\begin{cases}
\frac{\card{\pastobs_{[i]}}}{\card{\selected_{[i]}}} \text{, if } \card{\selected_{[i]}} > 0 \\
0\text{, otherwise}
\end{cases}
\end{equation}
where $\pastobs_{[i]}$ and $\selected_{[i]}$ denote the sets of recently observed and selected landmarks of appearance equivalence class $[\landmark_i]$ respectively.
Accordingly, we define our new landmark ranking function as:
\begin{equation}
\label{ranking_final}
f_{rank}(\landmark) \coloneqq p([\landmark] \mid \appearance) \text{, } \forall \landmark \in \alllandmarks
\end{equation}
In Figure~\ref{fig:iros2016_comp}, a comparison of our modified ranking function with the originally proposed formulation on the experimental set-up used in \cite{Burki2016Appearance-basedLocalization} is shown, ensuring that our modified formulation does not reveal regressive performance under these conditions. 
Further evaluation results demonstrating the merit of our new formulation are presented in Section \ref{subsec:evaluation_abl}.

%
%
%

\subsection{Offline Map-Summarization}
Whenever a $\richsession$ is added to the map, the total number of landmarks increases, and therewith also the size of the map.
%
%
%
%
We therefore apply the map summarization techniques proposed by Dymczyk et al. in \cite{Dymczyk2015KeepMaps} to keep the map size bounded at all times.

%
They suggest to reduce the number of landmarks by solving the following integer-based optimization problem:
\begin{align}
\text{minimize}\; \mathbf{q}^T\mathbf{x} + \lambda \mathbf{1}^T\boldsymbol{\zeta} \text{, subject to}
\end{align}
\begin{align}
\sum\nolimits_{i=1}^N \mathbf{x}_i &= n_{desired} \\
\mathbf{A}\mathbf{x} + \boldsymbol{\zeta} &\geq b\mathbf{1} \\
\boldsymbol{\zeta} &\in \lbrace\lbrace 0 \rbrace \cup \mathbb{Z}^+\rbrace^M.
\end{align}
Each landmark is assigned a corresponding binary switch variable $x_i \in \left\lbrace 0, 1 \right\rbrace$, indicating whether the landmark should be kept or removed.
The landmarks are selected based on the cost vector $\mathbf{q}$, estimated using the number of sessions a landmark was observed in and the total number of observations.
Additional constraints ensure some desired total number of landmarks ($n_{desired}$) to remain in the map, and a sufficient number of landmarks ($b$) visible from every vertex.
Matrix $\mathbf{A}$ encodes the vertex-landmark co-observability, while the slack variable $\boldsymbol{\zeta}$ allows to relax this constraint (at cost $\lambda$), ensuring a solution to the optimization problem can be found in all cases.
\section{Evaluation}
\label{sec:evaluation}
Our evaluation is structured into two parts as follows:
We first present our findings related to the map management process and offline summarization, thereby looking into how many $\richsessions$ are added over time, and how the degree of map summarization affects the localization performance.
In the second part, we show the performance of the online appearance-based landmark selection on the incrementally improved maps over time, focusing on a comparison between our modified, more generic ranking function proposed in Section~\ref{subsec:methodology_abl}, and the original ranking function proposed in~\cite{Burki2016Appearance-basedLocalization} under the influence of additional $\observationsessions$.

The data for the evaluation has been collected in two complementary real-world scenarios.
The first one covers weather and seasonal change over the coarse of a full year at day-time on an outdoor parking-lot, while the second one covers extreme lighting change from full day-light to complete night-time in a city environment. 
%
%
The sensor suite consists of four fish-eye cameras, one facing in each cardinal direction, running at $12.5$Hz, and wheel-odometry.
The images are scaled down to $640 \times 400$ pixels prior to processing.
Example images can be found in\cite{Burki2016Appearance-basedLocalization} and in the video contributions available online\footnote{\tiny{\url{https://youtu.be/TJMQCSHTIjU}}}\footnote{\tiny{\url{https://youtu.be/JL_5zMEQKYc}}}.
All computations have been performed on a consumer-grade laptop with an Intel i7 CPU.
Localization runs in real-time at $> 5$Hz.

%

\subsection{Map Update and Summarization}
\label{subsec:eval_map_update_and_summarization}
As a metric for assessing the quality of the generated maps over time, we employ translation RMS errors between the rough pose estimate from forward-propagated wheel-odometry and the refined pose after optimization.
%
%
In accordance with the results found in \cite{Burki2016Appearance-basedLocalization}, we omit the presentation of RMS errors in orientation, as they highly correlate with the translation errors and are of negligible magnitude in any case ($\ll 2^{\circ}$).
Note that the refined pose at each iteration is obtained from solving a vision-only non-linear least-squares optimization problem.
The resulting RMS error thus approximates the standard deviation of the localization along the trajectory.

In outdoor environments, the updated map must still be able to cover the range of appearance conditions represented by the incorporated $\richsessions$, even after summarization. 
The number of landmarks required to achieve this not only depends on the sensor setup and the spatial extent of the map, but also on the variance in appearance conditions encountered.
The City-Environment scenario covers extreme lighting changes from day-time to night-time, but the overall variance is still considerably smaller than in the year-long day-time Parking-Lot scenario. 
To guide our map update process described in Section~\ref{subsec:map_update} and Figure~\ref{fig:flowchart_map_update}, we have therefore chosen to perform map summarization with a maximum number of $75$k ($75'000$) and $150$k landmarks in the City-Environment and the Parking-Lot scenario respectively, and use a $10$cm threshold on the translation RMS error on these maps as a decision criterion to add the dataset at hand either as a $\richsession$, or as an $\observationsession$.
%
%
The choice of the $10$cm threshold is motivated by recent work (\cite{Muhlfellner2015SummaryLocalization}, \cite{Burki2016Appearance-basedLocalization}) suggesting this to be a reasonable and realistic upper bound for localization precision with the given sensor suite.
%

\begin{figure*}
\includegraphics[width=\textwidth]{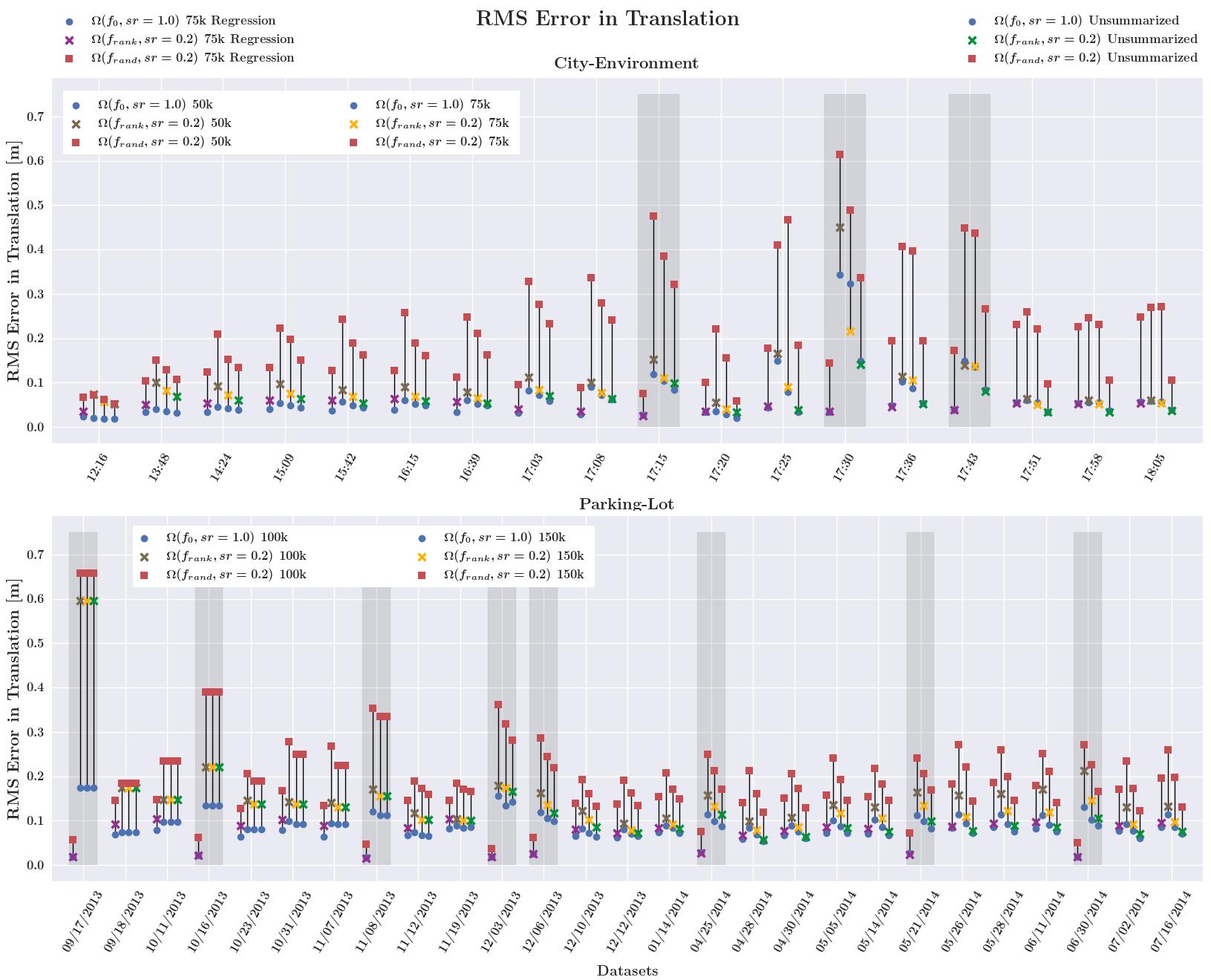}
\caption{Translation RMS error for all the datasets of both scenarios, differently summarized maps, and different ranking functions and selection policies.
The datasets marked in dark gray are added as $\richsessions$ while all other datasets are added as $\observationsessions$ instead.
The blue circles correspond to the precision achieved when using all landmarks for localization ($\Omega(f_0(), \selectionratio=1.0)$), whereas the red cubes correspond to $20\%$ random selection ($\Omega(f_{rand}(), \selectionratio=0.2)$, while the colored crosses represent the appearance-based selection ($\Omega(f_{rank}(), \selectionratio=0.2))$.
These results are shown for four different map configurations in each of the scenarios: The unsummarized map, two degrees of summarization, and with respect to the final summarized map (``Regression'').
\label{fig:zoo_fofa_precision}}
\vspace{-5mm}
\end{figure*}

The evolution of the localization performance resulting from this map update regime is shown in Figure~\ref{fig:zoo_fofa_precision}, where localization has been evaluated with the following three combinations of selection policies and ranking functions: a)\,\,${\Omega(f_{0}(),\,\selectionratio=1.0})$, using all candidate landmarks for localization, b)\,\,${\Omega(f_{rank}(),\,\selectionratio=0.2)}$, appearance-based landmark selection with a selection ratio of $20\%$, and c)\,\,${\Omega(f_{rand}(),\,\selectionratio=0.2)}$, the corresponding random selection.
As described in \ref{subsec:methodology_abl} and \cite{Burki2016Appearance-basedLocalization}, the selection policy ${\Omega(f(),\,\selectionratio = \alpha)}$ selects some fraction $\alpha$ of top-ranked landmarks from $\allcandidates_k$ which are then used for localization at the given iteration $k$.

%

%
%
%
In order to thoroughly assess the influence of the offline map summarization on the localization performance, we have further evaluated the latter against more strictly summarized maps (with $50k$ for the City-Environment, and $100k$ landmarks for the Parking-Lot environment respectively), as well as against indefinitely growing unsummarized maps.
In the City-Environment scenario, appearance conditions appear stable throughout the afternoon until the beginning of dusk shortly after $5$pm.
%
At $5$:$15$pm, $5$:$30$pm and $5$:$43$pm, additional $\richsessions$ are added, gradually expanding the appearance coverage of the map until, finally, night-time localization is feasible at $6$pm.

In contrast to that, in the Parking-Lot scenario, the appearance patterns are much less clear.
Already the initial map, built from the first dataset from August $20^{th}$, does not allow sufficient localization performance for the second dataset from September $17^{th}$.
In general, it seems to be necessary to have a $\richsession$ present for every month of the year.
Nevertheless, for the second half of the year, the map clearly shows converging tendencies, with only occasional datasets just barely above the $10$cm precision threshold, and the spread between reference localization using all landmarks, and the random selection decreasing.

\begin{figure}
\includegraphics[width=\textwidth/2]{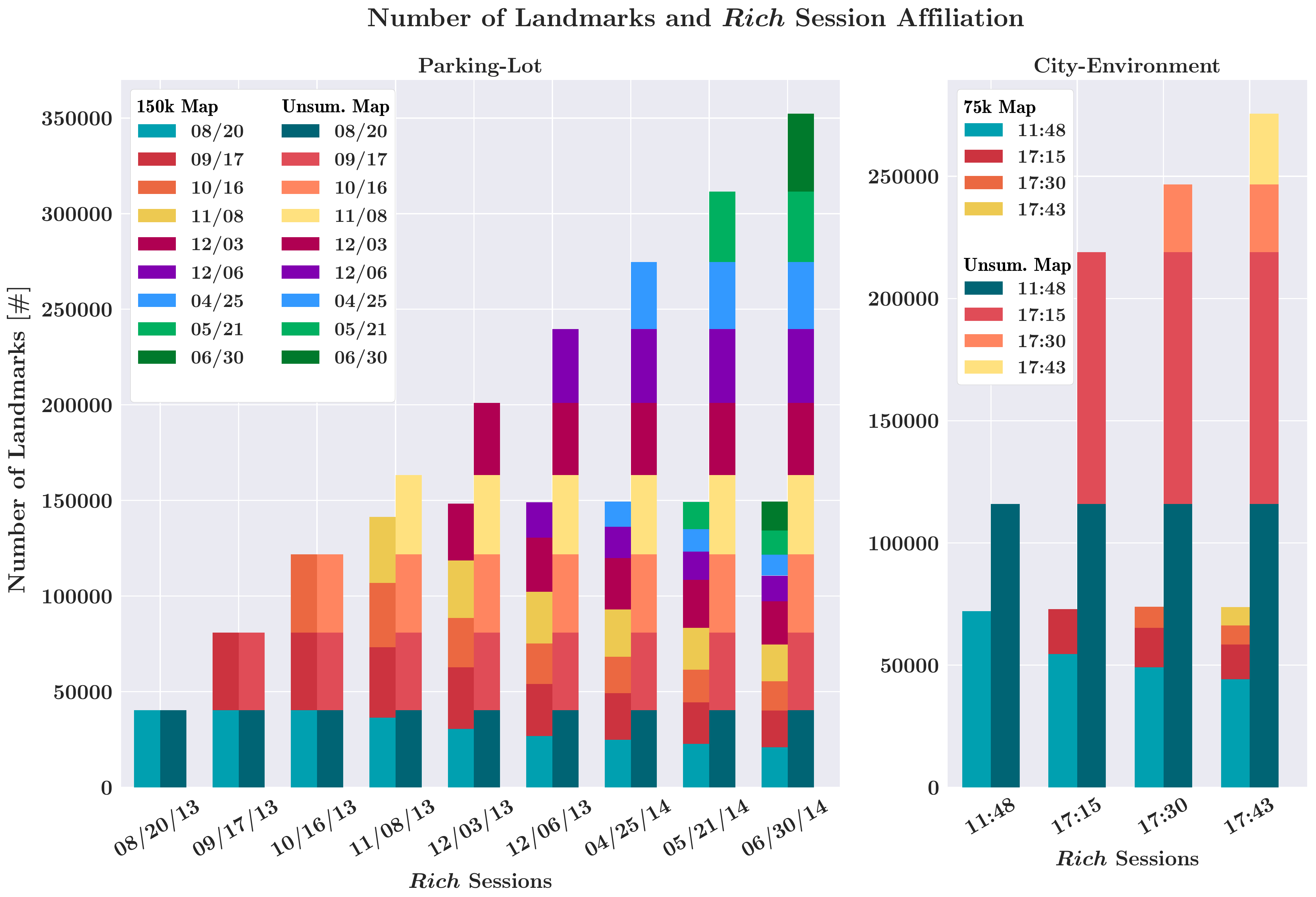}
\caption{The number of landmarks and their affiliation with the corresponding $\richsession$, both for the summarized and the unsummarized maps, is presented for the two evaluation scenarios. On the x-axis, every dataset which is added as a $\richsession$ is listed in chronological order, whereas the y-axis shows the absolute number of landmarks the summarized and unsummarized maps contain at this stage. The colors indicate how many landmarks belong to what $\richsession$. 
$\readability$
\label{fig:zoo_fofa_num_landmarks}}
\vspace{-6mm}
\end{figure}

Figure~\ref{fig:zoo_fofa_num_landmarks} further shows the number of landmarks associated with each $\richsession$ at each stage of the incremental map building process, both for the summarized map and the unsummarized map.
The $\richsessions$ added in the City-Environment scenario in dusk naturally contain considerably fewer landmarks, which is also reflected in both the summarized and the unsummarized map.
In contrast to that, the $\richsessions$ added in the Parking-Lot scenario all contain a similar number of landmarks, and summarization reduces already present sessions more or less equally as new sessions are added.


Ideally, the localization precision is preserved after map summarization.
If this is the case, the summarization only removes noisy landmarks from the map which are not re-observable under any of the encountered appearance conditions.
In the City-Environment scenario, the performance difference related to map summarization is best visible at night-time, where the unsummarized map shows significantly better performance in case of the random selection. 
However, with the appearance-based selection, almost the same precision is attainable as if all landmarks were used. 
This shows that the summarization algorithm successfully removes redundant and noisy landmarks while maintaining a good coverage over the different appearance conditions.
Similar results can be observed also for the Parking-Lot scenario.
The more $\richsessions$ are added, and hence the fewer landmarks of an individual $\richsession$ can be present in the summarized map, the larger the performance gap between the differently summarized and the unsummarized map becomes.
It can further be observed that for the $100$k map, the appearance-based localization cannot keep up with the performance compared to the $150$k map, indicating that for this scenario and this time span, more than $100$k landmarks are required in order to maintain sufficient appearance space coverage.

Since in these experiments we deliberately choose to build the maps incrementally and in chronological fashion, the performance evaluation of a certain dataset only uses the map available at that point in time.
To demonstrate that the summarization algorithm in fact creates maps that maintain usability across all previously encountered appearance conditions, we evaluate the performance of all datasets in retrospect using the final map created after having processed the last dataset in chronological order. 
The results of this ``regression'' test are shown in Figure~\ref{fig:zoo_fofa_precision}, with the corresponding map labelled with ``Regression''.
%
%
As can be seen, all datasets achieve at least as high a precision as if the map available by that time is used instead.
%
%
Note, however, that for all datasets added as $\richsessions$ this ``regression'' test in principle corresponds to self-localization.
Hence the artificially high precision in these cases.

\subsection{Appearance-Based Landmark Selection}
\label{subsec:evaluation_abl}
The goal of the appearance-based landmark selection is to achieve a high online localization performance with as few landmarks selected as possible.
This can best be evaluated by comparing respective selection ratios $\selectionratio$ with the corresponding observation ratios $r_{k}^{obs}$:

\begin{equation}
r_{k}^{obs} \coloneqq \frac{\card{\pastobs_k^{\Omega(f_{rank},\selectionratio=\alpha)}}}{\card{ \pastobs_k^{\Omega(f_{0}, \selectionratio=1.0)}}}
\label{eq:selection_percentage}
\end{equation}
The selection ratio $\selectionratio$  denotes the fraction of candidate landmarks used for localization at iteration $k$.
In contrast to that, the observation ratio $r_{k}^{obs}$ compares the number of observed landmarks, using a specified ranking function $f_{rank}()$ and selection ratio $\alpha$, to the number of observed landmarks when all candidate landmarks are used.

%
%

In \cite{Burki2016Appearance-basedLocalization}, it has been shown that with a pre-built map and selection ratios between $20$-$40\%$, a localization performance comparable to using all landmarks can be achieved. 
In contrast to that, in this paper, we aim at investigating how the relation between selection ratio and localization performance evolves in a scenario where datasets are chronologically processed and the map is built-up incrementally, with both $\richsessions$ and $\observationsessions$.
%
%
%
%

\begin{figure}
\includegraphics[width=\textwidth/2]{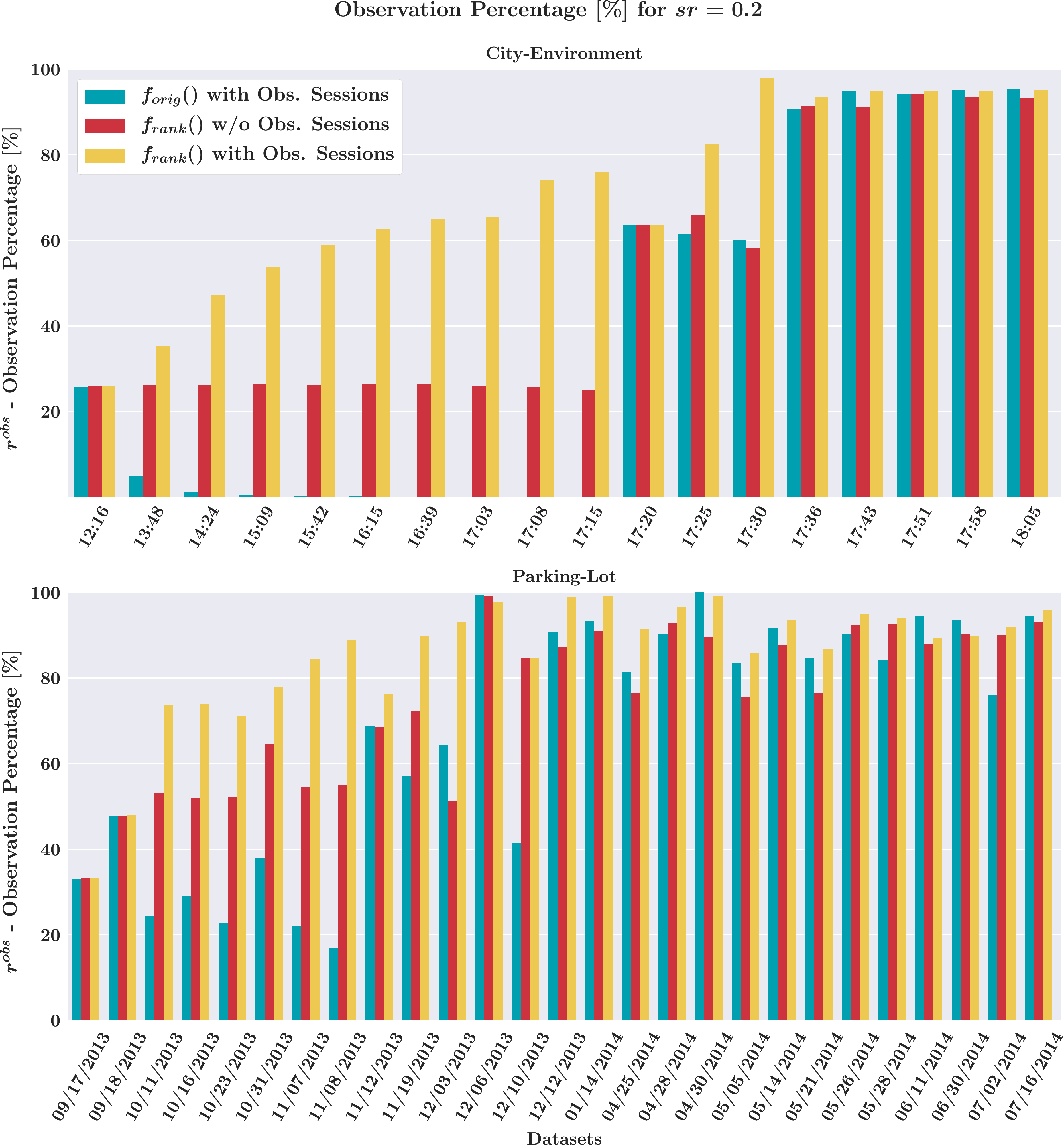}
\caption{The observation percentage $r^{obs}$ is shown for both scenarios, a selection ratio of $20\%$ and for the following three ranking functions and selection policies: 
a) $f_{orig}()$ with $\observationsessions$,
b) $f_{rank}()$ with $\observationsessions$, and
c) $f_{rank}()$ without $\observationsessions$.
Especially in the early stages where the map still contains only very few $\richsessions$, the $\observationsessions$ allow a significant boost of the observation percentage. 
\label{fig:zoo_fofa_ncm_bars}}
\vspace{-4mm}
\end{figure}

Figure~\ref{fig:zoo_fofa_ncm_bars} shows the observation percentage for a selection ratio of $20\%$ for the three cases of using the ranking function originally proposed in \cite{Burki2016Appearance-basedLocalization}, and using our new ranking function introduced in Section~\ref{subsec:methodology_abl} with and without $\observationsessions$.

In early stages of the map building with only few $\richsessions$ present in the map, the benefit of using the $\observationsessions$ is most pronounced. 
As soon as more $\richsessions$ become available though, the selection not using the $\observationsessions$ performs more and more similarly.
In case of the City-Environment scenario, this is due to the fact that towards night-time even a pure selection on the $5$:$30$pm and $5$:$43$pm $\richsessions$ allows achieving virtually $100\%$ observation percentage already.
In contrast to that, the appearance conditions in the Parking-Lot scenario are much more diverse and unrelated from one dataset to the next one. 
After having some number of $\richsessions$ present in the map, additional co-observability information in potentially only weakly related appearance conditions is only of minor or no help anymore.

\section{conclusions}
\label{sec:conclusions}
We have presented a complete map management process for a visual localization system tailored to long-term operations in resource constraint outdoor environments. 
Offline map summarization guarantees maps of bounded size at all times, while online localization with appearance-based landmark selection allows only transmitting and processing the map data required and useful under the current appearance condition.
%
With the incorporation of landmark co-observation statistics in the form of $\observationsessions$ in combination with a new formulation for the appearance-based landmark ranking function $f_{rank}()$, we have proposed a lightweight mechanism to improve the appearance-based landmark selection during online localization at negligible storage or computational costs.
%
%
An extensive evaluation in real-world conditions has shown that these additional $\observationsessions$ have the potential to significantly improve the landmark selection performance. 
However, their usefulness degrades as more and more $\richsessions$ are available in the map.
We have further evaluated the localization performance on the maps with different degrees of summarization resulting from the proposed map management paradigm, and shown that precise localization is possible over long time frames and across vastly different appearance conditions while keeping the map size bounded.


%

%
%

\addtolength{\textheight}{-10cm}   





\section*{ACKNOWLEDGMENT}

This project has received funding from the EU H2020 research project under grant agreement No 688652 and from the Swiss State Secretariat for Education, Research and Innovation (SERI) under contract number 15.0284.



{\small
\bibliographystyle{IEEEtran}
\bibliography{Mendeley_local}
}

\end{document}